\documentclass[10pt,twocolumn,letterpaper]{article}

\usepackage{iccv}
\usepackage{times}
\usepackage{epsfig}
\usepackage{graphicx}
\usepackage{amsmath}
\usepackage{amssymb}
\usepackage{xurl}
\usepackage{multiraw}

\iccvfinalcopy 

\def\iccvPaperID{****} 

\ificcvfinal\fi

\begin{document}

\title{StackMix and Blot Augmentations for Handwritten Text Recognition}

\author{
Alex Shonenkov\\
SBER AI \\
Sochi, Russian Federation \\
{\tt\small shonenkov@phystech.edu}
\and
Denis Karachev\\
OCRV\\
Sochi, Russian Federation \\
{\tt\small Denis.Karachev@OCRV.ru}
\and
Max Novopoltsev\\
SBER AI \\
Sochi, Russian Federation \\
{\tt\small MYNovopoltsev@sberbank.ru}
\and
Mark Potanin\\
SBER AI, MIPT \\
Moscow, Russian Federation \\
{\tt\small mark.potanin@phystech.edu}
\and
Denis Dimitrov\\
SBER AI, Lomonosov MSU \\
Moscow, Russian Federation \\
{\tt\small denis.dimitrov@math.msu.su}
}

\maketitle

\begin{abstract}

This paper proposes a handwritten text recognition (HTR) system that outperforms current state-of-the-art methods. 
The comparison was carried out on three of the most frequently used in HTR task datasets, namely Bentham, IAM, and Saint Gall. In addition, the results on two recently presented datasets, Peter the Great’s manuscripts and HKR Dataset, are provided.

The paper describes the architecture of the neural network and two ways of increasing the volume of training data: augmentation that simulates strikethrough text (HandWritten Blots) and a ”new text” generation method (StackMix), which proved to be very effective in HTR tasks. StackMix can also be applied to the standalone task of generating handwritten text based on printed text. 

\end{abstract}

\section{Introduction}

Handwriting text recognition is an vital task. Automation allows for a dramatical reduction in labor costs for processing correspondence and application forms and deciphering historical manuscripts. 

Handwritten text has several features because of both the inter- and intraclass variability - different instances of the same word written by different people may vary greatly, and the same character written by the same writer may look very different depending on the context when it was written. In addition to the above, historical records may contain flaws like ink drops and paper defects. Another problem with historical documents is the usually small amount of labeled data. 
Deciphering historical manuscripts is carried out with the help of rare specialists – historians and linguists; this significantly increases the deciphering cost, especially with many documents. Our approach involves having specialists label a relatively small number of documents, train the model, and apply it to the remaining data.

We propose a handwritten text recognition system, as well as two ways to increase the volume of training data: augmentation that simulates strikethrough text - HandWritten Blots and a "new text" generation method - StackMix.

\begin{figure}[t]
\begin{center}
   \includegraphics[width=0.9\linewidth]{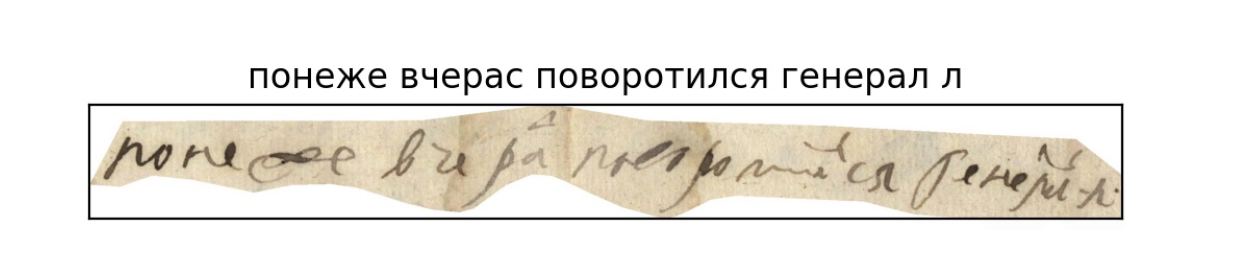}
\end{center}
   \caption{One line of text.}
\label{fig:one_line}
\end{figure}

The system was originally designed to decipher Peter the Great manuscripts that were first introduced at \cite{potanin2021digital} by using marked-up lines of the text as input. One line of text is 2 048 pixels width and 128 pixels height (Fig. \ref{fig:one_line}).

The remaining paper is organized as follows. Related works are presented in Section 2. Some of them are used for comparison with state-of-the-art models. Section 3 is devoted to describing our method. It includes information about neural network architecture and augmentations, and introduces StackMix. Section 4 describes datasets and metrics used in experiments. Section 5 provides a detailed description of the experiments and results. Section 6 concludes the paper.

\begin{figure*}
\begin{center}
    \includegraphics[width=0.9\linewidth]{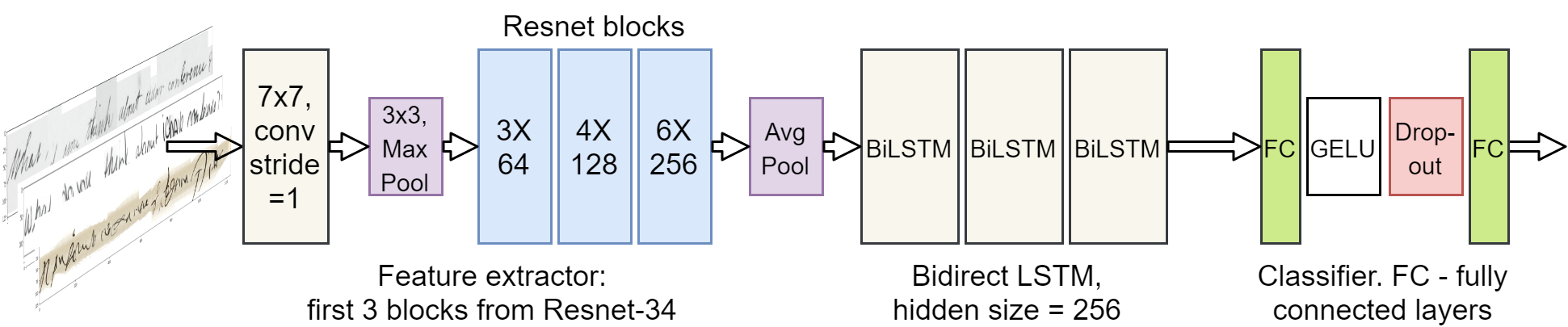}
\end{center}
   \caption{Neural network architecture.}
\label{fig:fig:nn}
\end{figure*}

\section{Related Work}

Early works on handwritten recognition problems suggest using a combination of hidden Markov models and RNNs \cite{bengio1999,bourlard1994} or algorithms based on conditional random fields \cite{lafferty2001}. The disadvantage of these approaches is the impossibility of end-to-end loss function optimization.
In 2006, a new approach was introduced - Connectionist Temporary Classification \cite{CTC}. The basic idea was to interpret the network outputs as a probability distribution over all possible label sequences, conditioned on a given input sequence. Given this distribution, an objective function can be derived that directly maximizes the probabilities of the correct labelings. Since the objective function is differentiable, the network can then be trained with standard backpropagation through time. New loss function were also introduced in this article - CTC loss. The ideas proposed in the article found a wide response among researchers, becoming the de-facto standard for handwritten recognition works \cite{gramCTC,HWRfew,de2019no}.

MDLSTM \cite{voigtlaender2016handwriting} networks use 2D-RNN which can deal with both axes of an input image. A simple model consists of several CNN and MDLSTM layers and using CTC loss provides excellent metrics for the IAM \cite{marti2002iam} dataset. However, MDLSTM models have some disadvantages like high computational costs and instability. In work \cite{de2019no}, the authors proposed an “example-packing” method. This eliminates the cost of padding for different-sized images. In works \cite{coquenet2020recurrence} and \cite{ingle2019scalable}, the authors try to eliminate recurrent layers in CNN-LSTM-CTC to decrease the number of parameters. Their Gated Fully Convolutional Networks shows relatively good results, even without a language model.

Another alternative to the RCNN-CTC approach is seq2seq models \cite{michael2019evaluating}. The encoder extracts features from the input, and the decoder with an attention mechanism emits the output sequentially. Common tricks may significantly improve the quality of HTR models. In \cite{aradillas2020boosting}, the authors investigated data augmentation and transfer learning for small historical datasets. 

OrigamiNet \cite{yousef2020origaminet} is a module that combines segmentation and text recognition tasks. It can be added to any line-level handwritten text recognition model to make it page-level.

We compared the proposed model with the models described above. The data for comparison is given in section 4.1, and the description of the metrics is given in section 4.2. We also found that different authors used various data partitions for train, test, and validation on the IAM dataset (see section 4.1). This made it difficult to correctly compare the models, so we are presenting our model results for the same data sets that were used in the original papers.



\section{Method}

Our method comprises three parts. Section 3.1 describes the modified Resnet neural network architecture we used. Sections 3.2 proposes a new augmentation method that simulates strikethrough text – Handwritten Blots. Finally, Section 3.3 describes how to significantly increase the amount of training data generating new text in the style of the current dataset (StackMix approach).

\subsection{Neural Network Architecture}

The neural network underlying the proposed system consists of three parts: a feature generator, a recurrent network to account for the order of the features, and the classifier that outputs the probability of each character.

\begin{figure}[t]
\begin{center}
    \includegraphics[width=0.9\linewidth]{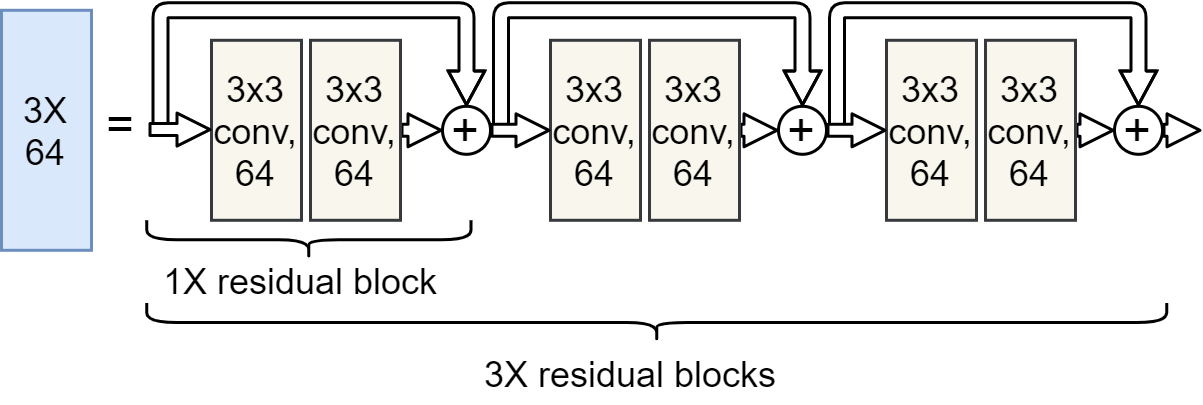}
\end{center}
   \caption{Resnet block architecture.}
\label{fig:fig:resnet}
\end{figure}

As a function generator, various network architectures were tested, and the final choice fell on Resnet (Fig. \ref{fig:fig:nn}). We took only three first blocks from Resnet-34 and replaced the stride parameter in the first layer with 1 to increase the "width" of the objects. One Resnet block (Figure \ref{fig:fig:resnet})  consisted of 3, 4, and 6 residual blocks with 64, 128, and 256 output layers, respectively.

After the features were extracted, they were averaged through the AdaptiveAvgPool2d layer and fed into the three BiLSTM layers to deal with feature sequences. As a final classifier, we use two fully connected layers with GELU and dropout between them. 

The results achieved using the described architecture without any additional modifications are shown in the "base" row of Table \ref{tab:all-results-table}. 


\subsection{Blot Augmentation}

The idea of augmentation appeared during the analysis of the Digital Peter dataset \cite{complink,potanin2021digital}. In the process of examining the dataset, we found examples of images in which some characters were crossed out and almost indistinguishable, but they were still present in the markup. Hense, the idea of using the Cutout augmentation \cite{devries2017improved} emerged, since it allows for overlaping of some elements of symbols or entire symbols, which makes the augmentation a bit like crossed-out symbols.



However, in the process of training the models, we got the idea of implementing such an algorithm that would allow for simulating the strikethrough characters as close as possible to the originals. Since we did not find the implementation of such algorithms in open sources, we created it ourselves.

To implement the strikethrough effect, we decided to use the Bezier curve construction algorithm, which in our case smoothed the curve transition between points.

The Bezier curve is a parametric curve and is a special case of the Bernstein polynomial. Finding basic polynomials of degree n are found by the formula:

\begin{equation}\label{eq4}
{b_{j,n}}=\left(\begin{array}{c}n\\j\end{array}\right)s^j\left(1-s\right)^{n-j}
\end{equation}

 Where $j=0,\dots ,n$. The definition of a curve as a linear combination is found as:

\begin{equation}\label{eq5}
B(s)=\sum_{j=0}^nb_{j,n}\cdot{v_j}
\end{equation}

Where $v_{j}$ is the point in the space, and $b_{j, n}$ define above. Since the sum of all polynomials must be equal to one, then.

\begin{equation}\label{eq6}
b_{0,n}+b_{0,1}+...+b_{n,n}=(s+(1-s))^n=1
\end{equation}

Where S is non-negative weights that sum to one. We found the implementation of the algorithm for constructing the Bezier curve in \cite{Hermes2017}.

\begin{figure}[ht]
\begin{center}
   \includegraphics[width=0.8\linewidth]{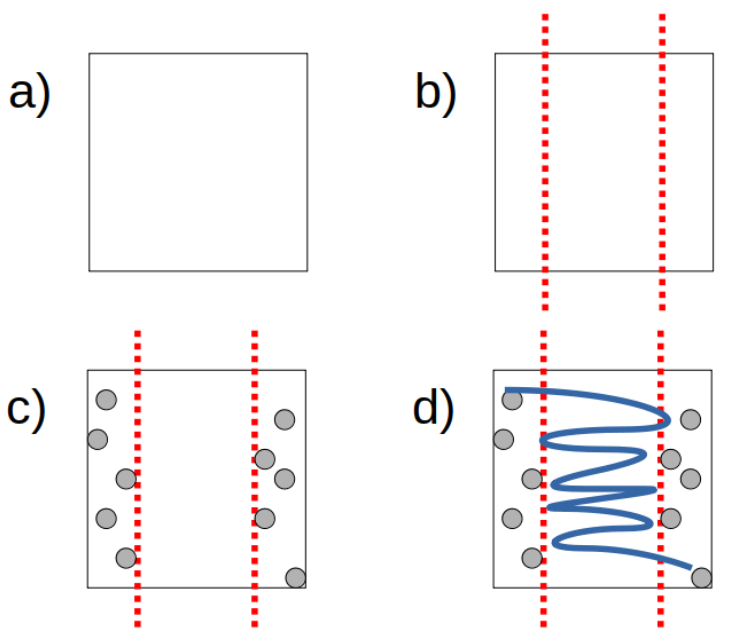}
\end{center}
   \caption{Graphic description of the algorithm. a) define the strikethrough area; b) define areas for generating points;  c) generate random points; d) draw a Bezier curve from the generated points}
\label{fig:hwb-algorithm-explanation}
\end{figure}

Next, we implemented our own algorithm that simulates strikethrough. The main steps were as follows:
\begin{enumerate}
    \item Determine the coordinates of the strikethrough area.
    \item Define areas for generating points to be used for drawing a Bezier curve.
    \item Generate points for the Bezier curve with the parameters of the intensity of the points and their coordinates to simulate the slope. Sometimes, a random point needs to be used several times for the loop to go a slightly further from the curve.
    \item Draw a Bezier curve with a specified transparency.
\end{enumerate}

\begin{figure}[t]
\begin{center}
   \includegraphics[width=0.8\linewidth]{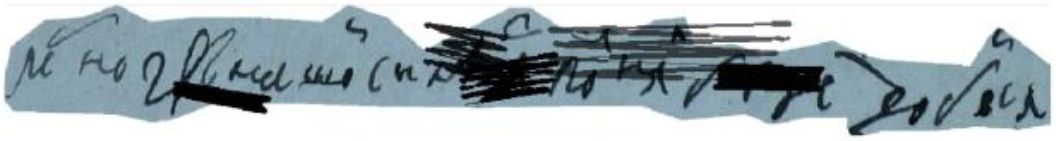}
\end{center}
   \caption{Sample image using faux strikethrough}
\label{fig:hwb-hwb-example}
\end{figure}

A graphical description of the algorithm is shown in Fig. \ref{fig:hwb-algorithm-explanation}. An example of a picture from a dataset using our strikethrough is provided in Fig. \ref{fig:hwb-hwb-example}

The implementation of this algorithm can be found here \cite{codelink}. We empirically selected the parameters for strikethrough (minh = 50, maxh = 100, minw = 10, maxw = 50, incline = 15, intensity = 0.9, transparency = 0.95, count = 1 \ldots 11, proba = 0.5) and tested it on different datasets.


The effect of the Blot augmentation on quality metrics is shown in the "blots" row of Table \ref{tab:all-results-table}, and in Figure \ref{fig:test_cer}. The obtained data suggests that handwritten blot augmentation makes a significant contribution to the quality of training models. Therefore, we recommend using it to train models in handwriting recognition problems.

\begin{figure*}
\begin{center}
    \includegraphics[width=1.0\linewidth]{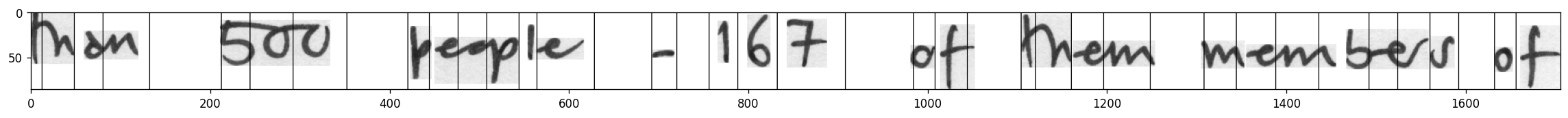}
    \label{fig:example1}
    \includegraphics[width=1.0\linewidth]{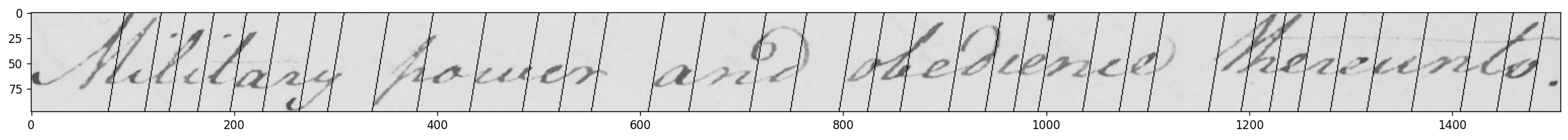}
    \label{fig:example2}
    \includegraphics[width=1.0\linewidth]{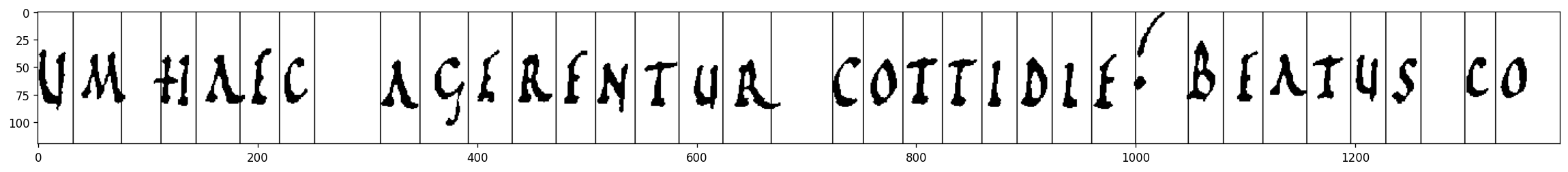}
    \label{fig:example3}
\end{center}
   \caption{Example images of symbol segmentation using semi supervised methods. Images have ids: IAM-a02-082-05, BenthamR0-072\_105\_002\_04\_05 and SaintGall-csg562-004-13 respectively.}
\label{fig:char-masks-examples}
\end{figure*}

\subsection{StackMix}
Improving the quality and stability of neural networks in image recognition tasks uses approaches that combine information from various objects of the training dataset, demonstrating effective performance and making the neural network more generalizable and resistant to new samples, despite not being previously presented to the neural network. For example, the CutMix approach \cite{yun2019cutmix} is an augmentation, in which parts of the images are cut from different samples and inserted into a new one, while the targets are mixed according to the proportions of the original parts of the images. A similar approach is used in SnapMix \cite{huang2020snapmix}, but the cutting of images is regulated by the neural network model using Class Activation Map. This approach allows for reducing the noise of the cut objects and selecting the most significant parts from the point of view of the neural network. Additionally, an interesting approach with mixing objects is presented in MixUp \cite{zhang2018mixup} and MWH \cite{yu2021mixup}, where images are overlapped on each other with a transparency coefficient and also mixes the targets with the proportion. Unfortunately, these methods cannot be applied to the optical character recognition and handwritten text recognition tasks, because mixing recursively dependent targets makes it very difficult to obtain a correct mapping of image and text. Therefore, we would like to introduce the StackMix approach, which allows for the improvement of the quality and stability of our neural network.

\begin{figure}
\begin{center}
    \includegraphics[width=1.0\linewidth]{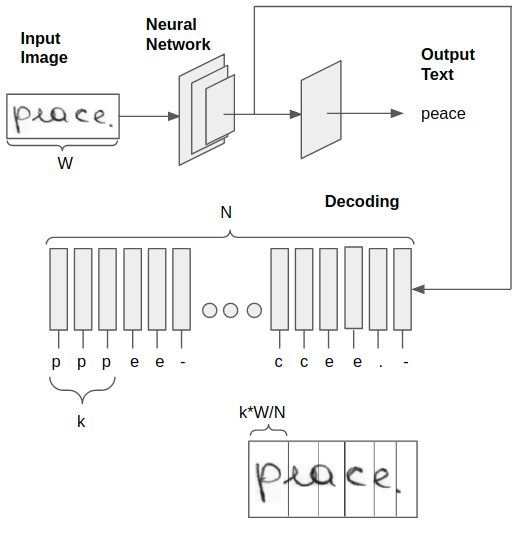}
\end{center}
   \caption{Post-processing scheme to get the boundaries of the symbols.}
\label{fig:char-masks}
\end{figure}

\medskip
\noindent 
\textbf{Weakly Supervised Symbol Segmentation.} First, to apply the proposed StackMix approach to the OCR task additional data markup that exactly marks the symbols boundaries is required. To achieve this an approach with automatic segmentation of training images into symbols using post-processing of a supervised pretrained neural network via CTC loss was used. The main idea was to connect the last layer of RNN (after applying SoftMax activation for every symbol from image features) and image width to get symbols' boundaries using only weakly supervised training without any manual markup (Fig. \ref{fig:char-masks}). For training neural network can be used base scheme without any augmentations and tricks. To get high quality marking of symbols' boundaries a sample from the training stage should be used. 


\medskip
\noindent 
\textbf{Corpus.} The StackMix approach also requires an external text corpus that has the same alphabet as the main dataset. Corpus does not require special marking and only contains allowed symbols. For the Digital Peter dataset, we used the texts of the XVII-XVIII centuries ($\sim$~3M lines). For the IAM and BenthamR0 datasets, we used cleaned texts from the Kaggle Competition "Jigsaw Unintended Bias in Toxicity Classification" \cite{jigsaw} ($\sim$~560k lines). For the HKR dataset, we used small portion of Russian texts from Wikimedia~\cite{ruwiki} ($\sim$~2.2M lines), and for the Saint Gall dataset, we used cleaned Latin corpus from Kaggle \cite{kaggle_latin_library} ($\sim$~180k lines) that used "The Latin Library" \cite{the_latin_library}.

\medskip
\noindent 
\textbf{StackMix Algorithm.} The input for the algorithm  is expected to be text from the external corpus, which creates a new image with this text using parts of images of the training dataset. The algorithm from nltk MWETokenizer \cite{nltkmwe} is used for tokenization. It processes tokenized text and merges multi-word expressions into single tokens. Collections of multi-word expressions are obtained from the training dataset, using symbol borders to connect parts of images and MWE tokens including spaces and punctuation marks. In this study, we used one random MWE tokenizer from six with a token dimension of no more than 3, 4, 5, 6, 7 and 8 with proba 0.05, 0.15, 0.2, 0.2, 0.2 and 0.2 respectively. After this each token wss matched with part of an image from the training data, and the pieces were stacked (hence the name "StackMix") together into a complete image, maintaining the correct order of the tokens.

\begin{figure}
\begin{center}
    \includegraphics[width=1.0\linewidth]{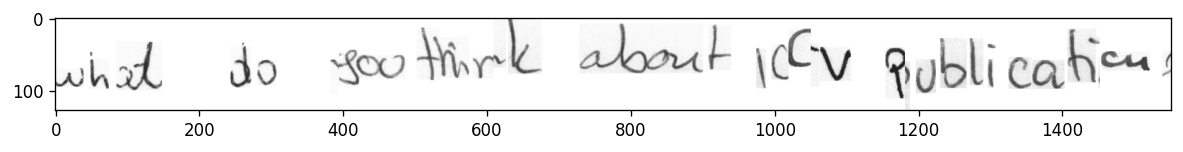}
    \label{fig:example1_c}
    \includegraphics[width=1.0\linewidth]{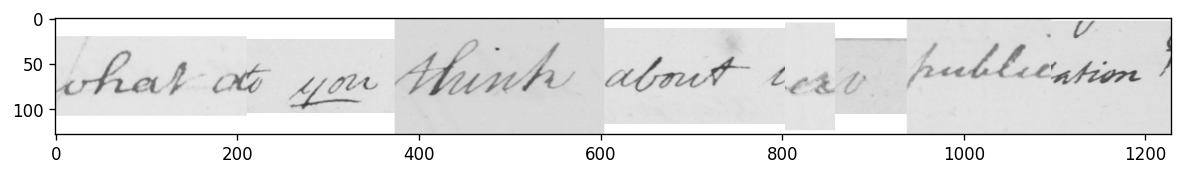}
    \label{fig:example2_c}
    \includegraphics[width=1.0\linewidth]{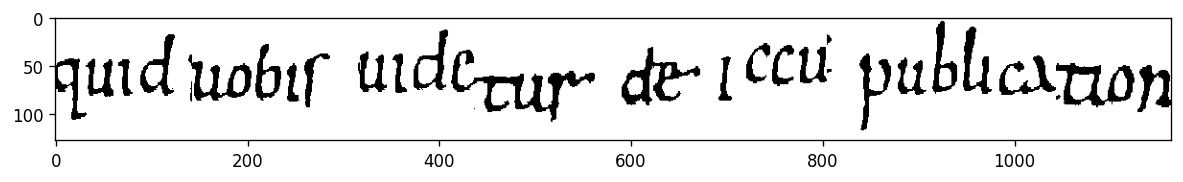}
    \label{fig:example3_c}
    \includegraphics[width=1.0\linewidth]{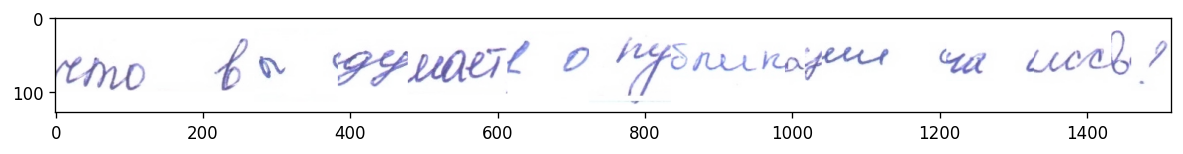}
    \label{fig:example4_c}
    \includegraphics[width=1.0\linewidth]{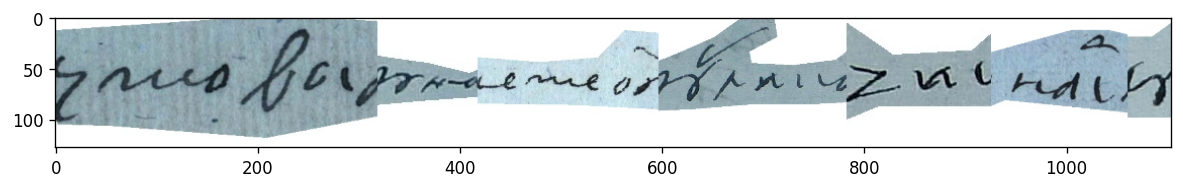}
    \label{fig:example5_c}
\end{center}
   \caption{Example images created using StackMix.}
\label{fig:stackmix-examples}
\end{figure}

Examples of the StackMix algorithm for various datasets are given in Figure \ref{fig:stackmix-examples} and the supplementary materials. Despite the visible places where tokens were glued together, the algorithm significantly increased the quality of recognition. The alignment and selection of samples to increase the realism of the generated string did not lead to an increase in metrics in our experiments. 

Nevertheless, after certain improvements, this algorithm may be used for the realistic generation of new documents. As an example in supplementary materials, we generated pages of texts from different sources. For all models, some paragraphs from the first chapter of a Harry Potter book was used. The results suggest that it is possible to generate different texts with different styles and fonts. For models of English language, we used the original Harry Potter book and for models that had Cyrillic symbols, the Russian version of the Harry Potter book was used.

\section{Benchmarks}

\subsection{Datasets}
Four different datasets were used in the experiments to prove state-of-the-art quality of our model.

\medskip
\noindent 
\textbf{Bentham} manuscripts refers to a large set of documents that were written by the renowned English philosopher and reformer Jeremy Bentham (1748-1832). Volunteers of the Transcribe Bentham\footnote{http://transcribe-bentham.ucl.ac.uk/td/Transcribe\_Bentham} initiative transcribed this collection. 
Currently, $>$ 6 000 documents or $>$ 25 000 pages have been transcribed using this public web platform.

For our experiments, we used the BenthamR0 dataset \cite{bentham} a part of the Bentham manuscripts.

\medskip
\noindent 
\textbf{The IAM} handwriting dataset contains forms of handwritten English text. It consists of 1 539 pages of scanned text from 657 different writers. This dataset is widely used for experiments in many papers. However, there is a big problem with uncertainty in data splitting for model training and evaluation. This issue was described in \cite{michael2019evaluating}. The IAM dataset has different train/val/test splits that are shown in Table \ref{tab:iam_split}. The problem is that none of them are labeled as a standard, so the IAM dataset split differs from paper to paper. However, results should be compared on the same split. 

In our experiments, we use the IAM-B\footnote{http://www.tbluche.com/resources.html} and IAM-D partitions. IAM-B was used to compare our model with others. IAM-D was a new partition inspired by the official page of  project\footnote{https://fki.tic.heia-fr.ch/databases/iam-handwriting-database}. This page contained "unknown","val1" and "val2" split labels. We added "unknown" samples to the train set, and combined "val1" and "val2" together.

IAM-B was chosen because many recently published papers used this partition. We used IAM-D because it provides more training samples.

We create a github repository \cite{iam_splits} with information and indexes corresponding to each IAM split in Table \ref{tab:iam_split}. It also contains links to papers that use these splits. We hope this helps researchers choose appropriate IAM partitions and make valid comparisons with other papers.


\begin{table}
  \label{tab:iam_split}
  \begin{center}
    \begin{tabular}{ |l|c|c|c| }
    \hline
    \textbf{Split} &  \textbf{Train} & \textbf{Val} & \textbf{Test} \\
    \hline
    IAM-A & 6161 & 966 & 2915 \\
    IAM-B & 6482 & 976 & 2915 \\
    IAM-C & 6161 & 940 & 1861 \\
    IAM-D & 9652 & 1840 & 1861 \\
    \hline
    \end{tabular}
    \end{center}
  \caption{IAM splits.}
\end{table}

\medskip
\noindent 
\textbf{Digital Peter} is a completely new dataset of Peter the Great's manuscripts \cite{potanin2021digital}. It consists of 9 694 images and text files corresponding to lines in historical documents. The open machine learning competition Digital Peter was held based on the considered dataset \cite{complink}. There are 6 237 lines in the training set, 1 527 lines in the validation set, and 1 930 lines in the testing set. 

\medskip
\noindent 
\textbf{HKR\_Dataset} \cite{nurseitov2020hkr} is a recently published dataset of modern Russian and Kazakh language. This database consists of $>$ 1 400 filled forms. It contains 64 943 lines and $>$ 715 699 symbols produced by about 200 different writers. Data are split in the following manner: 45 559 lines for a train set, 10 009 lines for validation, and 9 375 lines for test. This data splitting was found in github \cite{hkr_splitting_github} of the authors of the HKR\_Dataset, but these proportions of train/valid/test were slightly different that those of the original paper \cite{nurseitov2020hkr}. We assumed that the seed was not fixed in a script to get split, which was not a big problem for comparing results. Also, authors of the HKR\_Dataset used a good idea for splitting on Test1 and Test2 and demonstrated the problem of recognition text for new owners of the handwriting styles. The behavior of our approaches, separately in Test1 and Test2, can be found in the supplementary  materials.


\medskip
\noindent 
\textbf{Saint Gall} dataset contains handwritten historical manuscripts written in Latin that date back to the 9th century. It consists of 60 pages, 1 410 text lines and 11 597 words.

\subsection{Metrics}

\begin{figure*}
\begin{center}
    \includegraphics[width=0.9\linewidth]{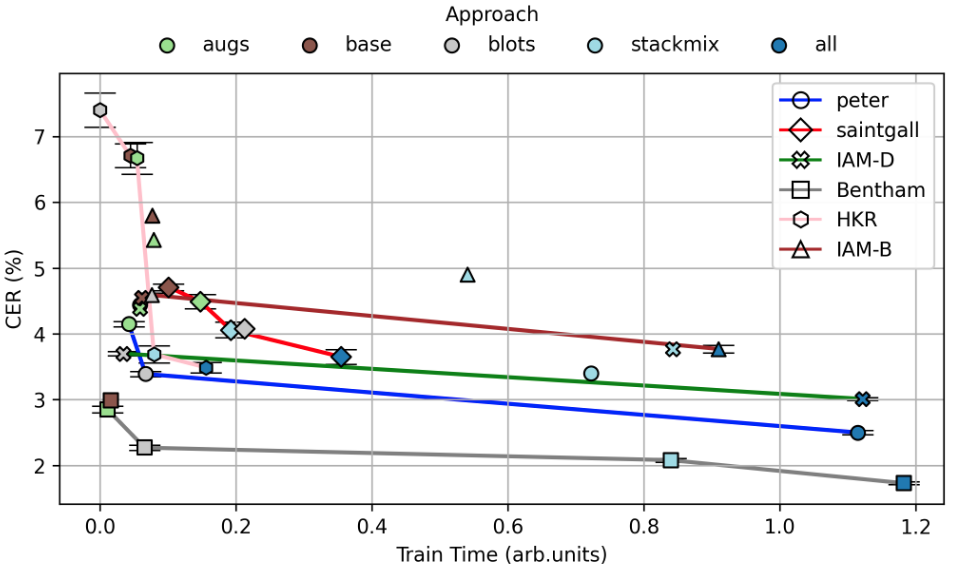}
\end{center}
   \caption{This graph compares the relative train time and CER results of experiments for different datasets and approaches. Arb. units for train time were obtained by the formula $T_{arb} = log_2(T/T_{min})$, where $T_{min} = 33.6$ ms is the minimum value of train time per one image. The colored lines represent obtained experiment points. Since the quality of approaches grows with increasing train time, some points have no line.}
\label{fig:test_cer}
\end{figure*}

We used character error rate (CER, $\%$), word error rate (WER, $\%$), string accuracy (ACC, $\%$) to measure the quality of the HTR models:

\begin{equation}\label{cer}
{\text{CER}} = \frac{\sum\limits_{i=1}^n {\sf dist}_c (pred_i, true_i)}{\sum\limits_{i=1}^n {\sf len}_c (true_i)},
\end{equation}
\begin{equation}\label{wer}
{\text{WER}} = \frac{\sum\limits_{i=1}^n {\sf dist}_w (pred_i,true_i)}{\sum\limits_{i=1}^n {\sf len}_w (true_i)},
\end{equation}
where ${\sf dist}_c$ and ${\sf dist}_w$ are the Levenshtein distances calculated in characters (including spaces) and words, respectively, and ${\sf len}_c$ and ${\sf len}_w$ are the length of the string in characters and words respectively.

\begin{equation}\label{accuracy}
{\text{ACC}} = \frac{\sum\limits_{i=1}^n [pred_i = true_i]}{n}.
\end{equation}

In the formulas \eqref{cer}, \eqref{wer}, and \eqref{accuracy}, $n$ is the size of the test sample, $pred_i$ is the string of characters that the model recognized in the $i$-th image, and $true_i$ is the true translation of the $i$-th image made by the expert.

\section{Experiments}

\subsection{Experimental Setup}

Each experiment was repeated five times. AdamW was used as an optimizer \cite{adamW} with OneCycleLR scheduler, starting at a learning rate of 0.001 down to 1e-08. The mean and standard deviation of each metric are given as the result. The use of augmentations allows for further metric improvements. In our experiments, we used StackMix ”on the fly” during training with a probability of 0.8. We added traditional augmentations of CLAHE \cite{Reza2004}, JpegCompression, Rotate, and our augmentation (simulation of crossed-out letters) - ”HandWritten Blots”. Different combinations of augmentations were grouped in our experiments:

\begin{itemize}
    \item "base" - experiments without augmentations, 300 epoch (HKR 100 epoch)
    \item "augs" - standart augmentations (CLAHE \cite{Reza2004}, JpegCompression, Rotate), 300 epoch (HKR 100 epoch)
    \item "blots" - using only our HandWrittenBlot augmentation, 500 epoch (HKR 150 epoch)
    \item "stackmix" - using only our Stackmix approach, 1~000 epoch (HKR 300 epoch)
    \item "all" - using all previous augmentations (augs + blots + stackmix), 1~000 epoch (HKR 300 epoch)
\end{itemize}

 Models with StackMix were trained during 1~000 epoch, but were not overfitted. We believe they should be trained more with bigger external text corpora. A comparison of our results for various datasets (IAM \cite{marti2002iam}, BenthamR0 \cite{bentham}, Digital Peter \cite{potanin2021digital}, HKR\_Dataset \cite{nurseitov2020hkr}, Saint Gall \cite{fischer2011transcription}) is presented in Table \ref{tab:all-results-table} and in Figure \ref{fig:test_cer}. 

Inference speed was measured as the average recognition speed for one line of text with batch size 16 and an image size of $128*2048$ pixels, using one Tesla-V100 GPU. The inference speed of different datasets are shown in Table \ref{tab:inference_speed} and inference speed of models with different augmentations is shown in Table \ref{tab:inference_speed2}.

\begin{table}
 
  \begin{center}
    \begin{tabular}{ |c|c| }
    \hline
    \textbf{Dataset} &  \textbf{Speed, img / sec}  \\
    \hline
    BenthamR0 & 86.2 ± 1.3 \\
    IAM & 94.9 ± 1.2 \\
    Digital Peter & 92.6 ± 2.7 \\
    HKR & 98.4 ± 1.3 \\
    Saint Gall & 86.3 ± 0.6 \\
    \hline
    \end{tabular}
    \end{center}
  \caption{ \label{tab:inference_speed}Inference speed for 1xV100 GPU, image size 128x2048.}
\end{table}

\begin{table}

    \begin{center}
    \begin{tabular}{ |c|c| }
    \hline
    \textbf{Experiment} &  \textbf{Speed, img / sec}  \\
    \hline
    base & 28.4 ± 0.9 \\
    augs & 28.5 ± 0.6 \\
    blots & 28.2 ± 1.4 \\
    stackmix & 21.0 ± 5.0 \\
    all & 17.7 ± 5.8 \\
    \hline
    \end{tabular}
    \end{center}
    \caption{\label{tab:inference_speed2}Training speed for 1xV100 GPU, image size 128x2048.}
\end{table}

\begin{table*}[ht]
\begin{center}
\begin{tabular}{|c|c|c|c|c|c|c|c|c|c|c|c|c|}
\hline
\multirow{3}{*}{ } 
& \multicolumn{6}{c|}{Datasets} \\
\cline{2-7}
& \multicolumn{3}{c|}{BenthamR0} &  \multicolumn{3}{c|}{IAM-B} \\
\cline{2-7}
& CER, \% & WER, \% & ACC, \% & CER, \% & WER, \% & ACC, \% \\
\hline
base & 2.99 ± 0.06 & 11.8 ± 0.3 & 52.1 ± 0.8 & 5.80 ± 0.08 & 18.9 ± 0.2 & 29.3 ± 0.4  \\
augs & 2.85 ± 0.05 & 11.3 ± 0.1 & 52.8 ± 1.0 & 5.43 ± 0.04 & 17.8 ± 0.1 & 30.7 ± 0.5  \\
blots & 2.27 ± 0.04 & 9.5 ± 0.2 & 57.0 ± 0.4 & 4.59 ± 0.03 & 15.0 ± 0.1 & 36.4 ± 0.5  \\
stackmix & 2.08 ± 0.03 & 9.0 ± 0.1 & 58.2 ± 0.8 & 4.90 ± 0.07 & 16.4 ± 0.2 & 35.2 ± 0.5 \\
all & \textbf{1.73 ± 0.02} & \textbf{7.9 ± 0.1} & \textbf{61.9 ± 1.1} & \textbf{3.77 ± 0.06} & \textbf{12.8 ± 0.2} & \textbf{43.6 ± 0.6} \\
\hline
\end{tabular}


\begin{tabular}{|c|c|c|c|c|c|c|c|c|c|c|c|c|}
\hline
\hline
\multirow{3}{*}{ } 
& \multicolumn{3}{c|}{Digital Peter} &  \multicolumn{3}{c|}{IAM-D} \\
\cline{2-7}
& CER, \% & WER, \% & ACC, \% & CER, \% & WER, \% & ACC, \% \\
\hline
base & 4.44 ± 0.02 & 24.3 ± 0.2 & 43.7 ± 0.5 & 4.55 ± 0.06 & 14.5 ± 0.2 & 35.5 ± 0.7 \\
augs & 4.15 ± 0.04 & 23.0 ± 0.2 & 45.7 ± 0.4 & 4.38 ± 0.04 & 14.0 ± 0.2 & 36.3 ± 0.6  \\
blots & 3.39 ± 0.04 & 19.3 ± 0.4 & 51.9 ± 0.4 & 3.70 ± 0.03 & 11.8 ± 0.1 & 42.3 ± 0.6  \\
stackmix & 3.40 ± 0.05 & 19.2 ± 0.3 & 51.6 ± 0.7 & 3.77 ± 0.04 & 12.3 ± 0.1 & 42.4 ± 0.8 \\
all & \textbf{2.50 ± 0.03} & \textbf{14.6 ± 0.2} & \textbf{60.8 ± 0.8} & \textbf{3.01 ± 0.02} & \textbf{9.8 ± 0.1} & \textbf{50.7 ± 0.3}  \\
\hline
\end{tabular}


\begin{tabular}{|c|c|c|c|c|c|c|c|c|c|c|c|c|}
\hline
\hline
\multirow{3}{*}{ } 
& \multicolumn{3}{c|}{HKR} &  \multicolumn{3}{c|}{Saint Gall} \\
\cline{2-7}
& CER, \% & WER, \% & ACC, \% & CER, \% & WER, \% & ACC, \% \\
\hline
base & 6.71 ± 0.18 & 22.5 ± 0.3 & 71.1 ± 0.5 & 4.71 ± 0.05 & 32.5 ± 0.3 & 2.2 ± 0.5 \\
augs & 6.67 ± 0.24 & 21.5 ± 0.3 & 72.2 ± 0.2 & 4.49 ± 0.11 & 31.3 ± 0.5 & 3.4 ± 0.7  \\
blots & 7.40 ± 0.26 & 23.0 ± 0.5 & 72.6 ± 0.4 & 4.08 ± 0.03 & 28.1 ± 0.2 & 4.6 ± 1.1 \\
stackmix & 3.69 ± 0.13 & 14.4 ± 0.4 & 80.0 ± 0.3 & 4.06 ± 0.12 & 28.8 ± 0.8 & 6.1 ± 0.8  \\
all & \textbf{3.49 ± 0.08} & \textbf{13.0 ± 0.3} & \textbf{82.0 ± 0.5} & \textbf{3.65 ± 0.11} & \textbf{26.2 ± 0.6} & \textbf{11.8 ± 2.0} \\
\hline
\end{tabular}

\end{center}
\caption{\label{tab:all-results-table} Results for all experiments}
\end{table*}

\subsection{Comparison with State-of-the-art}
In this section, we present the comparison with other models (Table \ref{tab:hwb-result-comparison}).

Our model outperforms other approaches on the BenthamR0, HKR, and IAM-D datasets. It reached 3.77\% of CER on the IAM-B dataset, which is very close to the best model published in 2016 that achieved 3.5\% CER. On the Saint Gall dataset, we achieved 5.56\% CER, which is very close to the current best solution of 5.26\% CER.

The IAM dataset has two versions because  papers use various data splits. We included IAM-B and IAM-D partitions in Table \ref{tab:hwb-result-comparison} to compare them with other models of the same split.

Authors of the paper \cite{de2020htr} provided open code for their model here \cite{htrflorlink}. We noticed that when evaluating the model, they lowered the characters in predicted and true strings. However, in our experiments, we did not convert the characters to lowercase. For real HTR tasks, it is not important to track the case of characters. In the supplementary materials, we provide the metrics of the models trained and evaluated on lowercase characters, and a comparison similar to the Table \ref{tab:hwb-result-comparison} on less widespread datasets.


\begin{table}[ht]
\begin{center}
\begin{tabular}{|c|c|c|}
\hline
\multirow{2}{*}{Model} & \multicolumn{2}{c|}{IAM-B}\\
\cline{2-3}
& CER, \% & WER, \% \\
\hline
\cite{voigtlaender2016handwriting} & \textbf{3.5} & \textbf{9.3} \\
\cite{yousef2020origaminet} & 4.7 & - \\
\cite{coquenet2020end} & 4.32 & 16.24 \\
\cite{coquenet2020recurrence} & 7.99 & 28.61\\
\cite{moysset20192d} & 7.73 & 25.22\\
\cite{wang2020decoupled} & 6.64 & 19.6\\
ours & 3.77$\pm$0.06  & 12.8$\pm$0.2\\
\hline
\hline
\multirow{2}{*}{} & \multicolumn{2}{c|}{IAM-D}\\
\cline{2-3}
\cite{abdallah2020attention}&7.8 & 25.5\\
ours & \textbf{3.01$\pm$0.02}  & \textbf{9.8$\pm$0.1}\\
\hline
\hline
\multirow{2}{*}{}& \multicolumn{2}{c|}{Digital Peter}\\
\cline{2-3}
\cite{complink}&10.5 & 44.4\\
ours & \textbf{2.50$\pm$0.03}  & \textbf{14.6$\pm$0.2}\\
\hline
\hline
\multirow{2}{*}{} & \multicolumn{2}{c|}{BenthamR0}\\
\cline{2-3}
\cite{abdallah2020attention} & 7.1 & 20.9 \\
\cite{de2020htr}& 3.98$\pm$0.06  & 9.8$\pm$0.14\\
ours & \textbf{1.73$\pm$0.02}  & \textbf{7.9$\pm$0.1}\\
\hline
\hline
\multirow{2}{*}{} & \multicolumn{2}{c|}{HKR}\\
\cline{2-3}
\cite{abdallah2020attention} & 4.5 & 19.2\\
ours & \textbf{3.49$\pm$0.08}  & \textbf{13.0$\pm$0.3}\\
\hline
\hline
\multirow{2}{*}{}& \multicolumn{2}{c|}{Saint Gall}\\
\cline{2-3}
\cite{abdallah2020attention} & 7.25 & 23.0\\
\cite{de2020htr}  & 5.26$\pm$0.03 & 21.14$\pm$0.13 \\
ours & \textbf{3.65$\pm$0.11}  & \textbf{26.2$\pm$0.6} \\
\hline
\end{tabular}
\end{center}
\caption{\label{tab:hwb-result-comparison} Comparison to other models, test set}
\end{table}

\section{Conclusion}
The neural network architecture presented in this article allows the problem of handwriting recognition of both modern and historical documents for various languages to be solved. The described augmentations - HandWritten Blots and StackMix - further improve the quality of recognition, demonstrating the best result among the currently known handwriting recognition systems.

The presented system can significantly increase the speed of deciphering historical documents. For example, it took a team of 10-15 historians about 3 months to decipher 662 pages of manuscripts from the Digital Peter dataset, what the organizers of the competition \cite{complink} wrote about. When working on the same dataset on a single Tesla V100, the average decryption speed was 95 lines/s or 380 pages/min, which is unattainable by historical scientists.

{\small
\bibliographystyle{ieee_fullname}
\bibliography{arxiv-version}
}

\end{document}